\documentclass[12pt]{article}
\usepackage{amsmath, amssymb, amsfonts, amscd, xspace, pifont, natbib}
\usepackage{epsfig, amsfonts, verbatim, multirow}
\usepackage{url, algorithm, algorithmic}
\usepackage{color}

\newcommand{\captionfonts}{\small}
\makeatletter  
\long\def\@makecaption#1#2{%
  \vskip\abovecaptionskip
  \sbox\@tempboxa{{\captionfonts #1: #2}}%
  \ifdim \wd\@tempboxa >\hsize
    {\captionfonts #1: #2\par}
  \else
    \hbox to\hsize{\hfil\box\@tempboxa\hfil}%
  \fi
  \vskip\belowcaptionskip}
\makeatother   

\newcommand{\V}[1]{\ensuremath{\boldsymbol{#1}}\xspace}

\newtheorem{theorem}{Theorem}
\newtheorem{lemma}{Lemma}

\newtheorem{proposition}{Proposition}

\begin{document}

\baselineskip = 1.7\baselineskip

\begin{titlepage}
\title{\Large \bf Sparse Ising Models with Covariates}
\author{Jie Cheng$^1$, Elizaveta Levina$^1$, Pei Wang$^2$ and Ji Zhu$^1$
 \\ \normalsize
$^1$Department of Statistics, University of Michigan, Ann Arbor, Michigan 48109, U.S.A.\\
 \normalsize
$^2$Fred Hutchinson Cancer Research Center, Seattle, Washington 98109, U.S.A.}

\maketitle
\bigskip
\bigskip

\abstract{There has been a lot of work fitting Ising models to multivariate binary data in order to understand the conditional dependency relationships between the variables.   However, additional covariates are frequently recorded together with the binary data, and may influence the dependence relationships.   Motivated by such a dataset on genomic instability collected from tumor samples of several types, we propose a sparse covariate dependent Ising model to study both the conditional dependency within the binary data and its relationship with the additional covariates.   This results in subject-specific Ising models, where the subject's covariates influence the strength of association between the genes.   As in all exploratory data analysis, interpretability of results is important, and we use $\ell_1$ penalties to induce sparsity in the fitted graphs and in the number of selected covariates.   Two algorithms to fit the model are proposed and compared on a set of simulated data, and asymptotic results are established.   The results on the tumor dataset and their biological significance are  discussed in detail.   }
	
\thispagestyle{empty}
\end{titlepage}

\section{Introduction}
Markov networks have been applied in a wide range of scientific and engineering problems to infer the local conditional dependency of the variables.  Examples include gene association studies \citep{Peng09, Wang11}, image processing \citep{Hassner80, Woods78}, and natural language processing \citep{Manning99}.  A pairwise Markov network can be represented by an undirected graph $G=(V,E)$, where $V$ is the node set representing the collection of random variables, and $E$ is the edge set where the existence of an edge is equivalent to the conditional dependency between the corresponding pair of variables, given the rest of the graph. 

Previous studies have focused on the case where an i.i.d.\ sample is drawn from an underlying Markov network, and the goal is to recover the graph structure, i.e., the edge set $E$, from the data.  Two types of graphical models have been studied extensively:  the multivariate Gaussian model for continuous data, and the Ising model \citep{Ising25} for binary data.   In the multivariate Gaussian case, the graph structure $E$ is completely specified by the off-diagonal elements of the inverse covariance matrix, also known as the precision matrix.  Therefore, estimating the edge set $E$ is equivalent to identifying the non-zero off-diagonal entries of the precision matrix.  Many papers on estimating the inverse covariance matrix have appeared in recent years, with a focus on the high-dimensional framework, for example, \citet{Meinshausen06, Yuan07, spice, banerjee06, Rocha08, Ravikumar08, Lam07,  Peng09, yuan10, Cai&Liu&Luo11}.    Most of these papers focus on penalized likelihood methods, and many establish asymptotic properties such as consistency and sparsistency.  Many have also proposed fast computational algorithms, the most popular of which is perhaps glasso by \cite{fht08}, which was recently improved further by \cite{Witten.etal2011} and \cite{mazumder&hastie2012}.

In the Ising model, the network structure can be identified from the coefficients of the interaction terms in the probability mass function.  The problem is, however, considerably more difficult due to the intractable normalizing constant, which makes the penalized likelihood methods popular for the Gaussian case extremely computationally  demanding.  \citet{Ravikumar10} proposed an approach in the spirit of \citet{Meinshausen06}'s work for the Gaussian case, fitting separate $\ell_1$-penalized logistic regressions for each node to infer the graph structure.   A pseudo-likelihood based algorithm was developed by \citet{Hofling09} and analyzed by \cite{ising}.

The existing literature mostly assumes that the data are an i.i.d.\ sample from one underlying graphical model, although the case of data sampled from several related graphical models on the same nodes has been studied both for the Gaussian and binary cases \cite{cgm, hising}.   However, in many real-life situations, the structure of the network may further depend on other extraneous factors available to us in the form of explanatory variables or covariates, which result in subject-specific graphical models.  For example, in genetic studies, deletion of tumor suppressor genes plays a crucial role in tumor initiation and development.  Since genes function through complicated regulatory relationships, it is of interest to characterize the associations among various deletion events in tumor samples. However, in practice we observe not only the deletion events, but also various clinical phenotypes for each subject, such as tumor category, mutation status, and so on.   These additional factors may influence the regulatory relationships, and thus should be included in the model.   Motivated by situations like this, here we propose a model for the conditional distribution of binary network data given covariates, which naturally incorporates covariate information into the Ising model, allowing the strength of the connection to depend on the covariates.   With high-dimensional data in mind, we impose  sparsity in the model, both in the network structure and in covariate effects.   This allows us to select important covariates that have influence on the network structure.  


There have been a few recent papers on graphical models that incorporate covariates, but they do so in ways quite different from ours.  \citet{Yin11} and \citet{Cai11} proposed to use conditional Gaussian graphical models to fit the eQTL (gene expression quantitative loci) data, but only the mean is modeled as a function of covariates, and the network remains fixed across different subjects.  \citet{Liu10} proposed a graph-valued regression, which partitions the covariate space and fits separate Gaussian graphical models for each region using glasso.  
This model does result in different networks for different subjects, but lacks interpretation of the relationship between covariates and the graphical model.  Further, there is a concern about stability, since the so built graphical models for nearby regions of the covariates are not necessarily similar.   In our model, covariates are incorporated directly into the conditional Ising model, which leads to straightforward interpretation and ``continuity'' of the graphs as a function of the covariates,  since in our model it is the strength of the edges rather than the edges themselves that change from subject to subject.

The rest of the paper is organized as follows.  In Section \ref{sec:model}, we describe the conditional Ising model with covariates, and two estimation procedures for fitting it.  Section \ref{sec:asymp} establishes asymptotic properties of the proposed estimation method.  We evaluate the performance of our method on simulated data in Section \ref{sec:sim}, and apply it to a dataset on genomic instability in breast cancer samples in Section \ref{sec:data}.  Section \ref{sec:discuss} concludes with a summary and discussion.

\section{Conditional Ising model with covariates}
\label{sec:model}

\subsection{Model set-up}
We start from a brief review of the Ising model, originally proposed in statistical physics by \cite{Ising25}. Let $\V{y} = (y_1, \ldots, y_q) \in \{0, 1\}^q$ denote a binary random vector.  The Ising model specifies the probability mass function $P_{\V{\theta}}(\V{y})$ as 
\begin{equation*}
P_{\V{\theta}}(\V{y}) = \frac{1}{Z(\V{\theta})}\textrm{exp}\left(\displaystyle \sum_{j} \theta_{jj}y_j + \sum_{k>j}\theta_{jk}y_jy_k\right),
\end{equation*}
where $\V{\theta} = (\theta_{11}, \theta_{12}, \ldots, \theta_{q-1q}, \theta_{qq})$ is a $q(q+1)/2$-dimensional parameter vector and $Z(\V \theta)$ is the partition function ensuring the $2^q$ probabilities summing up to 1.  Note that from now on we assume  $\theta_{jk}$ equals to $\theta_{kj}$ unless otherwise specified.  The Markov property is related to the parameter $\V\theta$ via
\begin{equation} \label{indwox}
	\theta_{jk} = 0 \Longleftrightarrow  y_j \perp y_k \parallel\ \V{y}_{\backslash(j,k)}, \ \ \ \ \forall  j\neq k,
\end{equation}
i.e., $y_j$ and $y_k$ are independent given all other $y$'s if and only if $\theta_{jk} = 0$.

Now suppose we have additional covariate information, and the data are a sample of $n$ i.i.d.\ points $\mathcal{D}_n = \{(\V{x}^1, \V{y}^1), \ldots, (\V{x}^n, \V{y}^n)\}$ with $\V{x}^i \in \mathbb{R}^p$ and $\V{y}^i \in \{0,1\}^q$.   We assume that given covariates $\V{x}$, the binary response $\V{y}$ follows the Ising distribution given by 
\begin{equation}
P(\V{y}|\V{x})  = \frac{1}{Z( \V{\theta}(\V{x}))} \ \exp\left(\displaystyle \sum_{j=1}^{q} \ \V{\theta}_{jj}(\V{x})y_j + \sum_{(j,k): 1\leq k < j \leq q} \V\theta_{jk}(\V{x})y_{j}y_k\right).
\end{equation}
We note that for any covariates $\V{x}^i$, the conditional Ising model is fully specified by the vector $\V\theta(\V{x}^i) = (\V\theta_{11}(\V{x}^i), \V\theta_{12}(\V{x}^i), \ldots, \V\theta_{q-1q}(\V{x}^i), \V\theta_{qq}(\V{x}^i))$, and by setting $\V\theta_{kj}(\V{x}) = \V\theta_{jk}(\V{x})$ for all $j > k$, the functions $\V{\theta}_{jk}(\V{x})$ can be connected to conditional log-odds in the following way,
\begin{equation} \label{loggeneral}
\log\left(\frac{P(y_j = 1 | \V{y}_{\backslash j},\V{x} )}{1-P(y_j = 1 | \V{y}_{\backslash j},\V{x}) }\right) = \V{\theta}_{jj}(\V{x}) + \displaystyle \sum_{k: k \neq j} \V\theta_{jk}(\V{x})y_{k},
\end{equation}
where, $\V{y}_{\backslash j} = (y_1,\ldots,y_{j-1},y_{j+1},\ldots,y_q)$.  Further, conditioning on $\V{y}_{\backslash\{j,k\}}$ being 0, we also have
\begin{equation*}
\log\left(\frac{P(y_j = 1,y_{k} = 1 |\  \V{y}_{\backslash\{j,k\}},\V{x})P(y_j = 0, y_{k} = 0 |\  \V{y}_{\backslash\{j,k\}},\V{x})}{P(y_j = 1,y_{k} = 0 |\  \V{y}_{\backslash\{j,k\}},\V{x})P(y_j = 0, y_{k} = 1 |\  \V{y}_{\backslash\{j,k\}},\V{x}) }  \right) = \V\theta_{jk}(\V{x}).
\end{equation*}
Similarly to \eqref{indwox}, this implies $y_j$ and $y_k$ are conditionally independent  given covariates $\V{x}$ and all other $y$'s if and only if $\V{\theta}_{jk}(\V{x}) =0$.

A natural way to model $\V{\theta}_{jk}(\V{x})$ is to parametrize it as a linear function of $\V{x}$.  Specifically, for $1 \leq j \leq k \leq q$, we let
\begin{eqnarray*}
\V \theta_{jk}(\V{x}) &=& \theta_{jk0} + \V \theta_{jk}^T\V{x},  \ \ \ \ \textrm{where} \ \ \V \theta_{jk}^T = (\theta_{jk1}, \ldots, \theta_{jkp}) \\
\V \theta_{jk}(\V{x}) &=& \V \theta_{kj}(\V{x}),  \ \ \ \ \ \ \ \ \ \ \ \forall j > k \ \ \  
\end{eqnarray*}
The model can be expressed in terms of the parameter vector $\V \theta = (\theta_{110}, \V{\theta}_{11}^T, \theta_{120}, \V{\theta}_{12}^T, \ldots, \theta_{qq0}, \V{\theta}_{qq}^T)$ as follows: 
\begin{equation} \label{model}
P_{\V \theta}(\V{y}|\V{x}) = \frac{1}{Z(\V{\theta}(\V{x}))} \exp\left(\displaystyle \sum_{j=1}^q(\theta_{jj0} + \V{\theta}_{jj}^T\V{x})y_j + \sum_{k>j} (\theta_{jk0}+\V\theta_{jk}^T\V{x})y_jy_{k}\right).
\end{equation}
Instead of  \eqref{loggeneral}, we now have the log-odds that depend on the covariates, through 
\begin{equation} \label{loglinear}
\log\left(\frac{P(y_j = 1| \V{y}_{\backslash j},\V{x} )}{1-P(y_j = 1| \V{y}_{\backslash j},\V{x} )}\right) = \theta_{jj0} + \V{\theta}_{jj}^T\V{x} + \displaystyle \sum_{k: k \neq j} (\theta_{jk0} +  \V\theta_{jk}^T\V{x})y_{k}.
\end{equation}

The choice of linear parametrization for $\V\theta_{jk}(\V{x})$ has several advantages. First, \eqref{loglinear} mirrors the logistic regression model when viewing the $x_\ell$'s, $y_k$'s and $x_{\ell}y_k$'s ($k\neq j$) as predictors. Thus the model has the same interpretation as the logistic regression model, where each parameter describes the size of the conditional contribution of that particular predictor.  Second, this parametrization has a straightforward relationship to the Markov network.  One can tell which edges exist and on which covariates they depend by simply looking at $\V{\theta}$. Specifically, the vector $(\theta_{jk0}, \V{\theta}^T_{jk})$ being zero implies that $y_k$ and $y_j$ are conditionally independent given any $\V{x}$ and the rest of $y_\ell$'s, and $\theta_{jk\ell}$ being zero implies that the conditional association between $y_j$ and $y_k$ does not depend on $x_\ell$.  Third, the continuity of linear functions ensures the similarity among the conditional models for similar covariates, which is a desirable property.  Finally, the linear formulation promises the convexity of the negative log-likelihood function,  allowing efficient algorithms for fitting the model discussed next.

\subsection{Fitting the model}
The probability model $P_{\V{\theta}}(\V{y}|\V{x})$ in \eqref{model} includes the partition function $Z(\V \theta(\V{x}))$, which requires summation of $2^q$ terms for each data point and  makes it intractable to directly maximize the joint conditional likelihood $\displaystyle \sum_{i=1}^n \log P_{\V{\theta}}(\V{y}^i | \V{x}^i)$.  However, \eqref{loglinear} suggests we can use logistic regression to estimate the parameters, an approach in the spirit of \citet{Ravikumar10}.  The idea is essentially to maximize the conditional log-likelihood of $y_j^i$ given $\V{y}_{\backslash j}^i$ and $\V{x}^i$ rather than the joint log-likelihood of $\V{y}^i$. 

Specifically, the negative conditional log-likelihood for $y_j$ can be written as follows
\begin{equation}
\ell_j (\V{\theta}; \mathcal{D}_n)  =  -\frac{1}{n}\sum_{i = 1}^n \log P(y_j^i | \V{x}^i, \V{y}_{\backslash j}^i) = - \frac{1}{n}\sum_{i = 1}^n\left(\log(1 + e^{\eta_j^i}) - y_j ^i \eta_j^i\right),
\label{eq-condlik}
\end{equation}
where 
\begin{equation*}
\eta_j^i = \log\left(\frac{P(y_j^i = 1| \V{y}^i_{\backslash j},\V{x}^i )}{1-P(y_j^i = 1| \V{y}^i_{\backslash j},\V{x}^i )}\right) = \V{\theta}_{jj}^T\V{x}^i + \sum_{k \neq j} (\theta_{jk0} + \V{\theta}_{jk}^T\V{x}^i)y_k^i.
\end{equation*}
Note that this conditional log-likelihood involves the parameter vector $\V{\theta}$ only through its subvector $\V{\theta}_{j} = (\theta_{j10}, \V{\theta}_{j1}^T, \ldots, \theta_{jq0}, \V{\theta}_{jq}^T) \in \mathbb{R}^{(p+1)q}$, thus we sometimes write $\ell_j (\V{\theta}_{j}; \mathcal{D}_n)$  when the rest of $\V{\theta}$ is not relevant. 

There are $(p+1)q(q+1)/2$ parameters to be estimated, so even for moderate $p$ and $q$ the dimension of $\V{\theta}$ can be large.  For example, with $p=10$ and $q=10$, the model has 605 parameters.  Thus there is a need to regularize $\V{\theta}$.  Empirical studies of networks as well as the need for interpretation suggest that a good estimate of $\V{\theta}$ should be sparse.  Thus we adopt the $\ell_1$ regularization to encourage sparsity, and propose two approaches to maximize the conditional likelihood \eqref{eq-condlik}.

\subsubsection*{Separate regularized logistic regressions}
The first approach is to estimate each $\V{\theta}_{j}$, $j=1, \ldots, q$ separately using the following criterion, 
\begin{equation*}
\displaystyle\min_{\V{\theta}_{j} \in \Bbb{R}^{(p + 1)q }} \ell_j (\V{\theta}_{j}; \mathcal{D}_n) + \lambda\|\V{\theta}_{j\backslash 0}\|_1,
\end{equation*}
where $\V{\theta}_{j\backslash 0} = \V{\theta}_{j} \backslash \{\theta_{jj0}\}$, that is, we do not penalize the intercept term $\theta_{jj0}$. 

In this approach, $\V{\theta}_{jk}$ and $\V{\theta}_{kj}$ are estimated from the $j$th and $k$th regressions, respectively, thus the symmetry $\hat{\V{\theta}}_{jk} = \hat{\V{\theta}}_{kj}$ is not guaranteed.  To enforce the symmetry in the final estimate, we post-process the estimates following \citet{Meinshausen06}, where the initial estimates are combined by comparing their magnitudes.  Specifically, let $\hat{\theta}_{jk\ell}$ denote the final estimate and $\hat{\theta}_{jk\ell}^0$ denote the initial estimate from the separate regularized logistic regressions.  Then for any $1\leq j <k\leq q$ and any $l = 0, \ldots, p$, we can use one of the two symmetrizing approaches: 
\begin{eqnarray*}
\textrm{separate-max:} &&\hat{\theta}_{jk\ell} = \hat{\theta}_{kj\ell} =   \hat{\theta}_{jk\ell}^{0} \mathbb{I}_{(|\hat{\theta}_{jk\ell}^{0}| > |\hat{\theta}_{kj\ell}^{0}|)} + \hat{\theta}_{kj\ell}^{0} \mathbb{I}_{(|\hat{\theta}_{jk\ell}^{0}| < |\hat{\theta}_{kj\ell}^{0}|)} \\
\textrm{separate-min:}  && \hat{\theta}_{jk\ell} = \hat{\theta}_{kj\ell} =   \hat{\theta}_{jk\ell}^{0} \mathbb{I}_{(|\hat{\theta}_{jk\ell}^{0}| < |\hat{\theta}_{kj\ell}^{0}|)} + \hat{\theta}_{kj\ell}^{0} \mathbb{I}_{(|\hat{\theta}_{jk\ell}^{0}| > |\hat{\theta}_{kj\ell}^{0}|)}
\end{eqnarray*}
The separate-min approach is always more conservative than separate-max in the sense that the former provides more zero estimates.  It turns out that when the sample size is small, the separate-min approach is often too conservative to effectively identify non-zero parameters.  More details are given in Section \ref{sec:sim}.

\subsubsection*{Joint regularized logistic regression}
The second approach is to estimate the entire vector $\V{\theta}$ simultaneously  instead of estimating the $\V{\theta}_{j}$'s separately, using the criterion,
\begin{equation*}
\displaystyle\min_{\V{\theta}\in \Bbb{R}^{(p + 1)q(q+1)/2}}  \displaystyle  \sum_{j=1}^q \ell_j (\V{\theta}; \mathcal{D}_n) + \lambda\|\V{\theta}_{\backslash 0}\|_1,
\end{equation*}
where $\V{\theta}_{\backslash 0} = \V{\theta} \backslash \{\theta_{110}, \theta_{220}, \ldots, \theta_{qq0}\}$. 
The joint approach criterion can be written as one large penalized logistic regression by careful rearranging of terms.  One obvious benefit of the joint approach is that $\hat{\V{\theta}}$ can be automatically symmetrized by treating $\V{\theta}_{jk}$ and $\V{\theta}_{kj}$ as the same during estimation.  The price, however, is that it is computationally much less efficient than the separate approach.  

To fit the model using either the separate or the joint approach, we adopt the coordinate shooting algorithm in \citet{Fu98}, where we update one parameter at a time and iterate until convergence.  The implementation is similar to the glmnet algorithm of \citet{Friedman&Hastie&Tibshirani08_glmnet}, and we omit the details here.

\section{Asymptotics: consistency of model selection}
\label{sec:asymp}
In this section we present the model selection consistency property for the separate regularized logistic regression.  Results for the joint approach can be derived in the same fashion by treating the joint regression as a single large logistic regression.  The spirit of the proof is similar to \citet{Ravikumar10}, but since their model does not include covariates $\V{x}$, both our assumptions and conclusions are different.

In this analysis, we treat the covariates $\V{x}_i$'s as random vectors.
With a slight change of notation, we now use $\V{\theta}_j$ to denote  $\V{\theta}_{j\backslash 0}$, dropping the intercept which is irrelevant for model selection.  The true parameter is denoted by $\V{\theta}^*$.  Without loss of generality we assume that $\theta_{jj0}^* = 0$, and we also assume that  $\hat{\theta}_{jj0}=0$. 

First, we introduce additional notation to be used throughout this section. Let 
\begin{eqnarray}
\label{def_IU}
\V{I}_j^* &=& \Bbb{E}_{\theta^*}( \nabla^2 \log P_{\V{\theta}}(y_j | \V{x}, \V{y}_{\backslash j})) \\
                   &=& \Bbb{E}_{\theta^*}\left(p_j(1 - p_j)(\V{x} \otimes \V{y}_{\backslash j})(\V{x} \otimes \V{y}_{\backslash j})^T\right) \ \ (\textrm{Information matrix})\\
\V{U}_j^* &=&\Bbb{E}_{\theta^*}\left((\V{x} \otimes \V{y}_{\backslash j})(\V{x} \otimes \V{y}_{\backslash j})^T\right)
\end{eqnarray}
where 
\begin{eqnarray*}
p_j &=& p_j(\V{x}, \V{y}_{\backslash j}) = P_{\V{\theta}^*}(y_j = 1 | \V{x}, \V{y}_{\backslash j}) \ , \\
\V{x} \otimes \V{y}_{\backslash j} &=& (1, x_1, \ldots, x_p)^T \otimes (y_1, \ldots, y_{j-1}, 1, y_{j+1}, \ldots, y_{q})^T \backslash \{1\} \ .
\end{eqnarray*}
Let $\mathcal{S}_j$ denote the index set of the non-zero elements of $\V{\theta}_j^*$, and let  $\V{I}_{\mathcal{S}_j\mathcal{S}_j}^*$ be the submatrix of $\V{I}^*_j$ indexed by $\mathcal{S}_j$. Similarly defined are $\V{I}_{\mathcal{S}_j^c\mathcal{S}_j}$ and $\V{I}_{\mathcal{S}_j^c\mathcal{S}_j^c}$, where $\mathcal{S}_j^c$ is the compliment set of $\mathcal{S}_j$.  Moreover, for any matrix $A$, let 
$\| A  \|_{\infty} = \max_i \sum_j |A_{ij}|$ be the matrix $L_\infty$ norm, and let $\Lambda_{\min}(A)$ and $\Lambda_{\max}(A)$ be the minimum and maximum eigenvalues of $A$, respectively.

For our main results to hold, we make the following two assumptions for all $q$ logistic regressions.
\begin{description}
\item[A1] There exists a constant $\alpha \in (0,1]$, such that
\begin{equation*}
\|\V{I}_{\mathcal{S}_j^c\mathcal{S}_j}^* \left(\V{I}_{\mathcal{S}_j\mathcal{S}_j}^*\right)^{-1}\|_{\infty} \leq (1 - \alpha)\ .
\end{equation*}

\item[A2] There exist constants $\Delta_{\min} > 0$ and $\Delta_{\max} > 0$, such that 
\begin{eqnarray*}
\Lambda_{\min}\left( \V{I}_{\mathcal{S}_j\mathcal{S}_j}^*\right) &\geq& \Delta_{\min} \\
\displaystyle \Lambda_{\max}(\V{U}_j^*) &\leq& \Delta_{\max}
\end{eqnarray*}
\end{description}
These assumptions  bound the correlation among the effective covariates, and the amount of dependence between the group of effective covariates and the rest.  Under these assumptions, we have the following result:
\begin{theorem} \label{thm:main}
For any $j = 1, \ldots, q$,  let $\V{\hat{\theta}}_j$ be a solution of the problem
\begin{equation}\label{minimization}
\displaystyle \min_{\V{ \theta}_j} \ \ -\ell_j(\V \theta_j; \mathcal{D}_n) + \lambda_n \|\V {\theta}_j\|_1.
\end{equation}
Assume $\V{A1}$ and $\V{A2}$ hold for $\V{I}_j^*$ and $\V{U}_j^*$, and further assume that  for some $ \delta > 0$ 
\begin{equation} \label{condition_xbound}
P(\|\V{x}\|_\infty \geq M) \leq \exp(-M^\delta), \ \  \ \textrm{ for all } \ M \geq M_0 >0, 
\end{equation} 
Let $d = \max_j\|\mathcal{S}_j\|_0$ and $C > 0$ a constant independent of $(n, p, q)$.  If 
\begin{eqnarray}
\label{condition_M}M_n &\geq& (C\lambda_n^2n)^{\frac{1}{1+\delta}}  \ , \\
\label{condition_lambda}\lambda_n &\geq& CM_n\sqrt{\frac{\log p+\log q}{n}} \ ,\\
\label{condition_n}n &\geq& CM_n^2d^3(\log p+\log q) \ , 
\end{eqnarray}
the following hold with probability at least $1 - \exp^{ -C(\lambda_n^2n)^{\delta^*}}$ ($\delta^*$ is a constant in (0, 1)),
\begin{enumerate}
\item Uniqueness: $\hat{\V{\theta}}_j$ is the unique optimal solution for any $j \in \{1, \ldots, q\}$. 
\item $\ell_2$ consistency: $\|\V{\hat{\theta}}_j - \V{\theta}_j^*\|_2 \leq 5\lambda_n\sqrt{d} / \Delta_{\min}$ for any $j \in \{1, \ldots, q\}$ 
\item Sign consistency: $\V{\hat{\theta}}_j$ correctly identifies all the zeros in $\V{\theta}_j^*$ for any $j \in \{1, \ldots, q\}$; moreover, $\V{\hat{\theta}}_j$ identifies the correct sign of non-zeros in $\V{\theta}_j^*$ whose absolute value is at least $10 \lambda_n\sqrt{d} / \Delta_{\min}$. 
\end{enumerate}
\end{theorem}

Theorem \ref{thm:main} establishes the consistency of model selection allowing both of the dimensions $p(n)$ and $q(n)$ to grow to infinity with $n$.  The extra condition, which requires the distribution of $\V{x}$ to have a fast decay on large values, was not in \citet{Ravikumar10} as the paper does not consider covariates.  The new condition is, however, quite general; for example, it is satisfied by the Gaussian distribution and all categorical covariates.  The proof of the theorem can be found in the Appendix.

\section{Empirical performance evaluation}
\label{sec:sim}
In this section, we present three sets of simulation studies designed to test the model selection performance of our methods.  We vary different aspects of the model, including sparsity, signal strength and proportion of relevant covariates.  The results are presented in the form of ROC curves, where the rate of estimated true non-zero parameters (sensitivity) is plotted against the rate of estimated false non-zero parameters (1-specificity) across a fine grid  of the regularization parameter.  Each curve is smoothed over 20 replications.

The data generation scheme is as follows.  For each simulation, we fix the dimension of the covariates $p$, the dimension of the response $q$, the sample size $n$ and a graph structure $E$ in the form of a $q \times q$ adjacency matrix (randomly generated scale-free networks \citep{Barabasi1999}.   For any $(j,k)$, $1\leq j\leq k \leq q$, $(\theta_{jk0}, \V{\theta}^T_{jk})$ consists of $(p+1)$ independently generated and selected from three possible values:  $\beta > 0$ (with probability $\rho/2$),  $- \beta$ (with probability $\rho/2$), and 0 (with probability $1-\rho$).   An exception is made for the intercept terms $\theta_{jj0}$, where $\rho$ is always set to 1.  Covariates $\V{x}^{i}$'s are generated independently from the multivariate Gaussian distribution $N_p(0,I_p)$. Given each $\V{x}^i$ and $\V \theta$, we use Gibbs sampling to generate the $\V{y}^i$, where we iteratively generate a sequence of $y_j^i$'s $(j = 1, \ldots q)$ from a Bernoulli distribution with probability $P_{\V{\theta}}(y_j^i = 1 | \V{y}_{\backslash j}^i, \V{x}^i)$ and take the last value of the sequence when a stopping criterion is satisfied.  

We compared three estimation methods:  the separate-min method, the separate-max method and the joint method.  Our simulation results indicate that performance of the separate-min method is substantially inferior to that of the separate-max method in almost all cases (results omitted for lack of space).  Thus we only present results for the separate-max and the joint methods in this section.

\subsection{Effect of sparsity}

First, we investigate how the selection performance is affected by the sparsity of the true model.  The sparsity of $\V \theta$ can be controlled by two factors: the number of edges in $E$, denoted by $n_E$, and the average proportion of effective covariates for each edge, $\rho$.  We fix the dimensions $q=10$, $p=20$ and the sample size $n =200$, and set the signal size to $\beta = 4$.  Under this setting, the total number of parameters is 1155. The sparsity parameter $n_E$ takes values in the set $\{10,20,30\}$, and $\rho$ takes values in $\{0.2, 0.5, 0.8\}$. 

\begin{figure}[h]
\hspace{-0.8in}
\includegraphics[width = 8in, height = 5in]{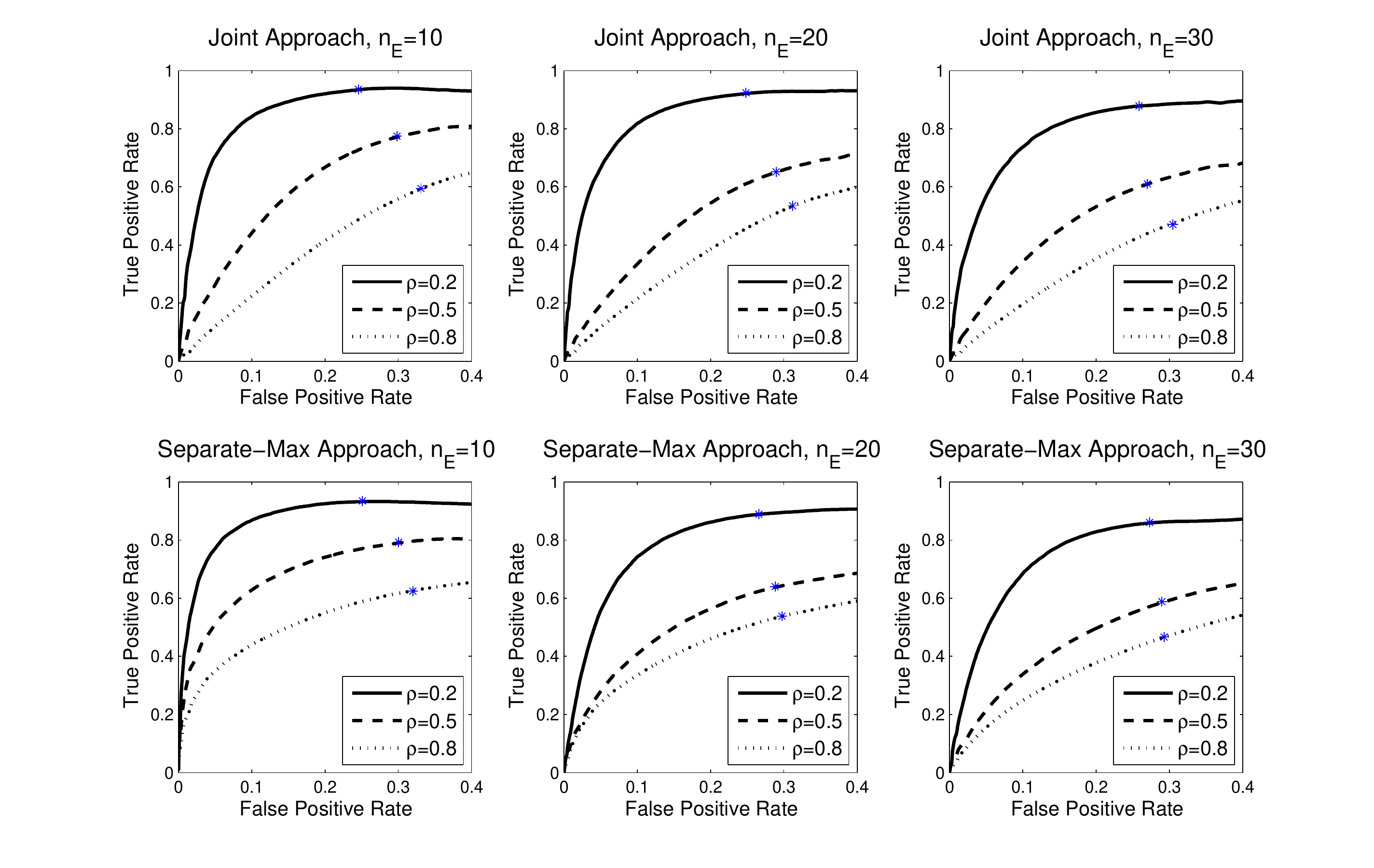}
\caption{ROC curves for varying levels of sparsity, as measured by the number of edges ($n_E$) and expected proportion of non-zero covariates ($\rho$).  The star on each curve corresponds to an optimal value of $\lambda$ selected on an independent validation set. }
\label{fig:sparsity}
\end{figure}

The resulting ROC curves are shown in Figure \ref{fig:sparsity}. The first row shows the results of the joint approach and the second row of the separate-max approach. As the true model becomes less sparse, the performance of both the joint and the separate methods deteriorates, since sparse models have the smallest effective number of parameters to estimate and benefit the most from penalization.    Note that the model selection performance seems to depend on the total number of non-zero parameters ($(q+n_E)(p+1)\rho$), not just on the number of edges ($n_E$). For example, both approaches perform better in case $n_E = 20, \rho = 0.2$ than $n_E = 10, \rho = 0.5$, even though the former has a more complicated network structure. Comparing the separate-max method and the joint method, we observe that the two methods are quite comparable, with the joint method being slightly less sensitive to increasing the number of edges.

Note that the ``$\ast$'' point on each curve represents the average sensitivity and (1-specificity) over the replications based on an ``optimal'' $\lambda$, selected by maximizing the conditional log-likelihood 
on an independent validation dataset of the same size as the training data. 

\subsection{Effect of signal size}
\begin{figure}[h]
\hspace{-0.5in}
\includegraphics[width = 7.2in, height = 3in]{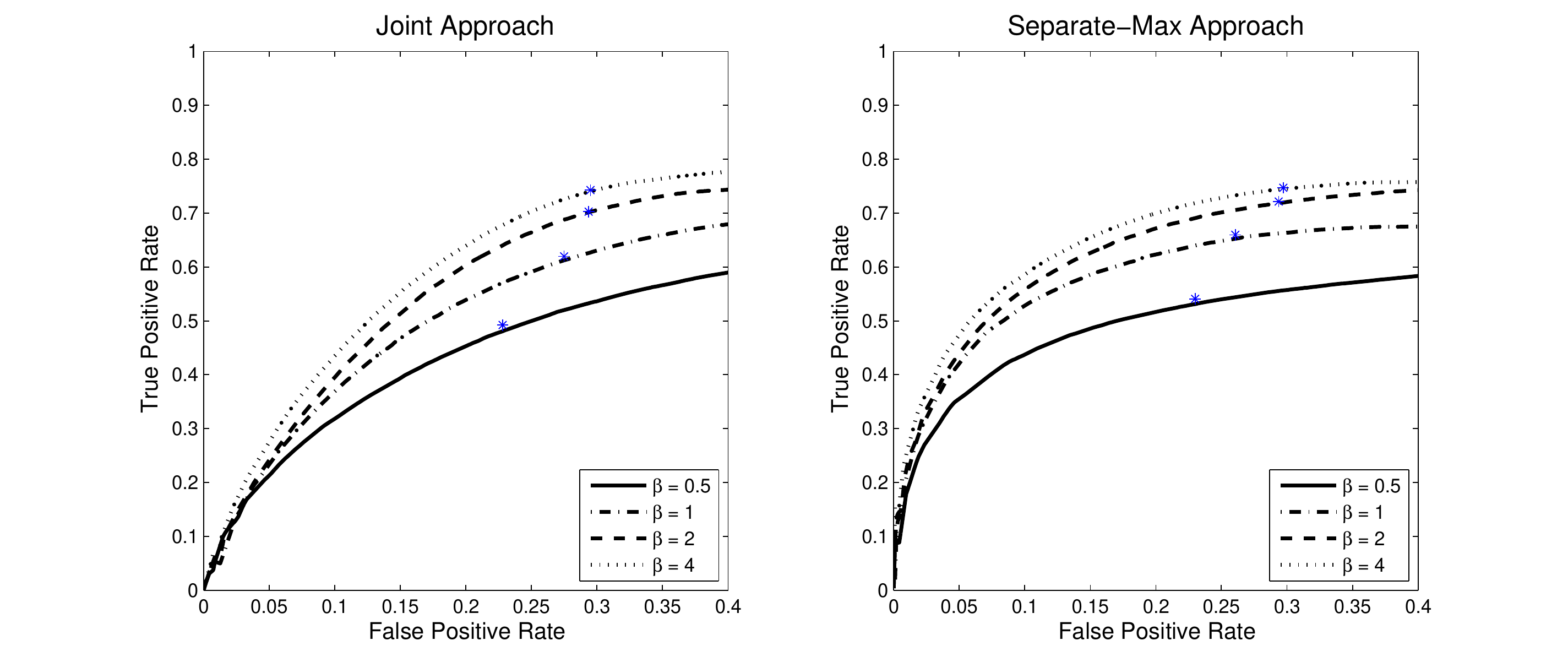}
\caption{ROC curves for varying levels of signal strength, as measured by the parameter $\beta$.  The star on each curve corresponds to an optimal value of $\lambda$ selected on an independent validation set. }
\label{fig:signal}
\end{figure}

Second, we assess the effect of signal size. The dimensions are set to be the same as in the previous simulation, that is, $q = 10$, $p = 20$ and $n = 200$, and underlying network is the same.  
The expected proportion of effective covariates for each edge is $\rho = 0.5$. The signal strength parameter $\beta$ takes values in the set $\{0.5, 1, 2, 4, 8, 16\}$.  For each setting, the non-zero entries of the parameter vectors $\V {\theta}$ are at the same positions with the same signs, only differing in magnitude. The resulting ROC curves are shown in Figure \ref{fig:signal}. 

As the signal strength $\beta$ increases, both the separate and the joint methods show improved selection performance, but the improvement levels off eventually.   Both methods achieve almost the same ``optimal'' sensitivity and specificity (the '$\ast$' point), with the separate-max method performing better overall.

\subsection{Effect of noise covariates}
In the last set of simulations, we study how the model selection performance is affected by adding extra uninformative covariates.  At the same time, we also investigate the effect of the number of relevant covariates $p_{\mbox{true}}$ and the sample size $n$. The dimension of the response is fixed to be $q=10$ and the network structure remains the same as in the previous simulation. We take $p_{\mbox{true}} \in \{10, 20\}$ and $n \in \{200, 500\}$. For each combination, we first fit the model on the original data and then on augmented data with extra uninformative covariates added. The total number of covariates $p_{\mbox{total}} \in \{p_{\mbox{true}}, 50, 200\}$. The non-zero parameters are generated the same way as before with $\beta=4$ and $\rho=0.5$. With the changes in $p_{\mbox{total}}$, the total number of non-zero parameters remains fixed for each value of $p_{\mbox{true}}$, while the total number of zeros is increasing. 

\begin{figure}[h!]
\hspace{-0.3in}
\includegraphics[width = 7in, height = 6.6in]{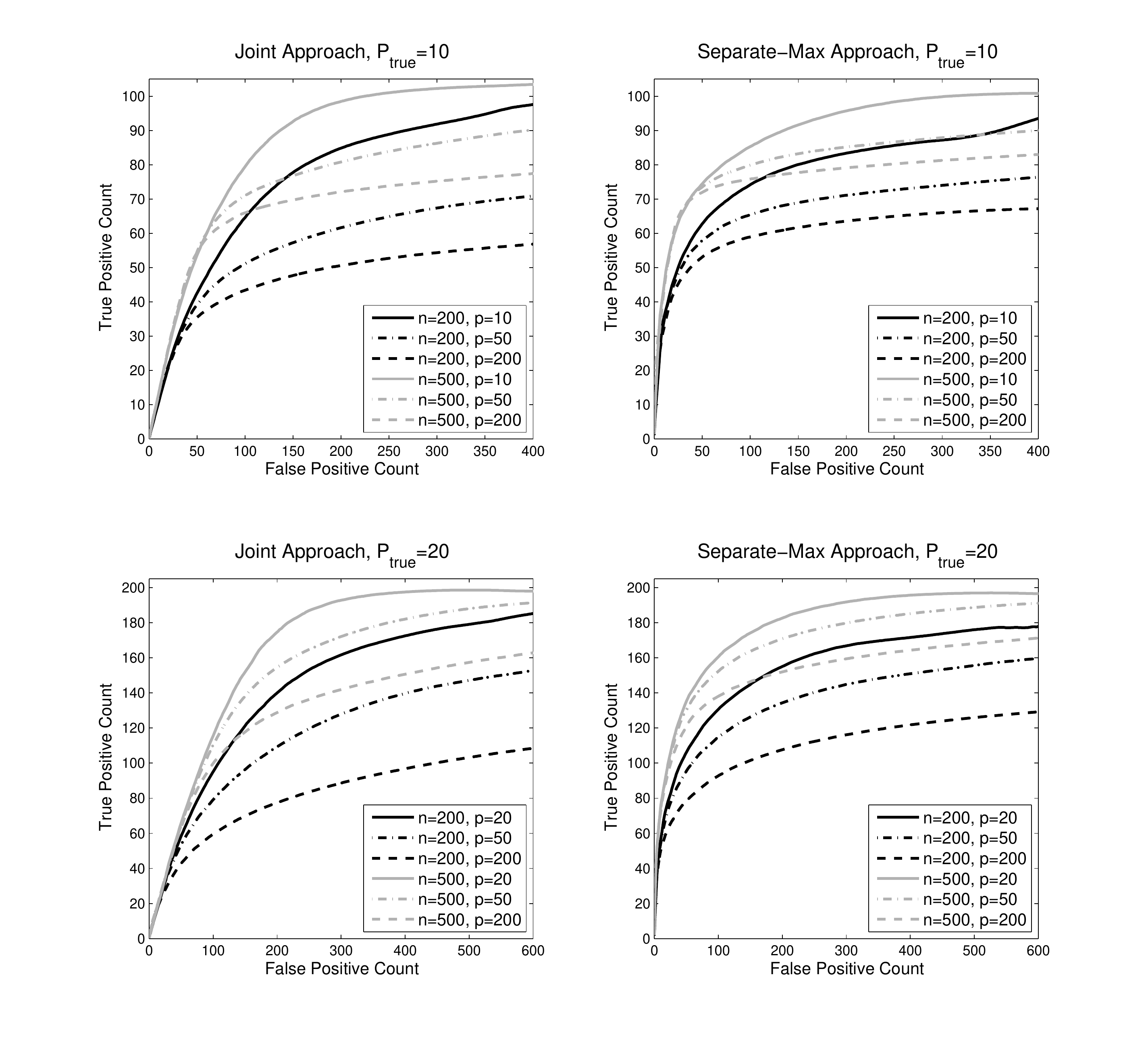}
\vspace{-0.5in}
\caption{ROC curves for varying dimension, number of noise covariates, and sample size. }.
\label{fig:noise}
\end{figure}

To make the results more comparable across setting, we plot the counts rather than rates of true positives and false positives.  The resulting curves are shown in Figure \ref{fig:noise}.   Generally, performance improves when the sample size grows and deteriorates when the number of noise covariates increases, particularly with a smaller sample size.   The separate-max method dominates the joint method under these settings, but the difference is not large. 

\section{Application to tumor suppressor genes study}
\label{sec:data}

In breast cancer, deletion of tumor suppressor genes plays a crucial role in tumor initiation and development. Since genes function through complicated regulatory relationships, it is of interest to characterize the associations among various deletion events in tumor samples, and at the same time to investigate how these association patterns may vary across different tumor subtypes or stages. 

Our data set includes DNA copy number profiles from cDNA microarray experiments on 143 breast cancer specimens \citep{Bergamaschi06}. Among them, 88 samples are from a cohort of Norwegian patients with locally advanced (T3/T4 and/or N2) breast cancer, receiving doxorubicin (Doxo) or 5 fluorouracil/mitomycin C (FUMI) neoadjuvant therapy \citep{Geisler03}. The samples were collected before the therapy.  The other 55 are from another cohort of Norwegian patients from a population-based series \citep{Zhao04}.  Each copy number profile reports the DNA amounts of 39,632 probes in the sample. The array data was preprocessed and copy number gain/loss events were inferred as described in \cite{Bergamaschi06}. To reduce the spatial correlation in the data, we bin the probes by cytogenetic bands (cytobands). For each sample, we define the deletion status of a cytoband to be 1 if at least three probes in this cytoband show copy number loss. 430 cytobands covered by these probes show deletion frequencies greater than 10\% in this group of patients, and they were retained for the subsequent analysis.  The average deletion rate for all the 430 cytobands in 143 samples is $19.59\%$. Our goal is to uncover the association among these cytoband-deletion events and how the association patterns may change with different clinical characteristics, including TP53 mutation status (a binary variable), estrogen receptors (ER) status (a binary variable), and tumor stage (an ordinal variable taking values in $\{1,2,3,4\}$).      
     
For our analysis, denote the array data by $\V{y}_{143\times 430}$, where $ y^i_j$ indicates the deletion status of the $j^{th}$ cytoband in the $i^{th}$ sample.   Let $\V{x}^i$ denote the covariate vector containing the three clinical phenotypes of the $i^{th}$ sample, and $\V{x}_l$ the $l$th covariate vector. We first standardize the covariate matrix $\V{x}_{143\times 3}$ and then fit our Ising model with covariates with the separate-max fitting method.   We then apply stability selection \citep{Meinshausen10} to infer the stable set of important covariates for each pairwise conditional association.    Specifically, we repeatedly fit the model 100 times on subsamples containing half the data selected randomly without replacement.  For each tuning parameter $\lambda$ from a fixed grid of values, we record the frequency of $\hat{\theta}_{jkl}$ being non-zero respectively for each covariate $\V{x}_l$, $l = 0,1,2,3$ on all pairs of $(j,k)$, $1 \leq j < k \leq 430$, and denote it by $f_{jkl}(\lambda)$. Note that $\V{x}_0$ corresponds to the main effect interaction between a pair of $\V{y}_j$'s and does not involve any covariates.     Then we use $f_{jkl}^* = \max_{\lambda} f_{jkl}(\lambda)$ as a measure of importance  of covariate $\V{x}_l$ for the edge $(j,k)$.  Finally, for each covariate $\V{x}_j$, we rank the edges based on the selection frequencies $\{f_{jkl}^*: 1<j\leq k<q\}$.     At the top of the list are the edges that depend on $\V{x}_j$ most heavily.    We are primarily interested in the pairs of genes belonging to different chromosomes, as the interaction between genes located on the same chromosome is more likely explained by strong local  dependency. The results are shown in Table 1, where the rank list of the edges depending on different covariates are recorded. The first two columns of each covariate related columns are the node names and the third columns record the selection frequency.  


There are 332 inter-chromosome interactions (between cytobands from different chromosomes) with selection probabilities at least 0.5. Among these, 39 interactions change with the TP53 status; 12 change with the ER status; and another 12 change with the tumor grade (see details in Table \ref{tab:results}). These results can be used by biologists to generate hypotheses and design relevant experiments to better understand the molecular mechanism of breast cancer. 

The most frequently selected pairwise conditional association is between deletion on cytoband 4q31.3 and deletion on 18q23 (94\% selection frequency). Cytoband 4q31.3 harbors the tumor suppressor candidate gene SCFFbw7, which works cooperatively with gene TP53 to restrain cyclin E-associated genome instability \citep{Minella07}. Previous studies also support the existence of putative tumor suppressor loci at cytoband 18q23 distal to the known tumor suppressor genes SMAD4, SMAD2 and DCC \citep{Huang95, Lassus01}.    Thus the association between the deletion events on these two cytobands is intriguing.

Another interesting finding is that the association between deletion on cytoband 9q22.3 region and cytoband 12p13.31 appears to be stronger in the TP53 positive group than in the TP53 negative group.  A variety of chromosomal aberrations at 9p22.3 have been found in different malignancies including breast cancer \citep{Mitelman97}.  This region contains several putative tumor suppressor genes (TSG), including DNA-damage repair genes like FANCC and XPA.  Alterations in these TSGs have been reported to be associated with poor patient survival \citep{Sinha08}. On the other hand, cytoband 12p13.31 harbors another TSG, namely ING4 (inhibitor of growth family member 4), whose protein binds TP53 and contributes to the TP53-dependent regulatory pathway. A recent study also suggests involvement of ING4 deletion in the pathogenesis of HER2-positive breast cancer.   In light of these previous findings, it is interesting that our analysis also found the association between the deletion events of 9p22.3 and 12p13.31, as well as the changing pattern of the association under different TP53 status. This result suggests potential cooperative roles for multiple tumor suppressor genes in cancer initiation and progression.

\begin{table}[htbp]\small
\centering
  \caption{Frequency-based ranked list of covariate-dependent inter-chromosomal interactions}
\label{tab:results}
    \begin{tabular}{lll|lll | lll }
    \hline
    \multicolumn{3}{c|}{Main effect} & \multicolumn{3}{c|}{TP53 mutation status} & \multicolumn{3}{c}{ER status} \\  \hline
    Gene1 & Gene2 & Freq  & Gene1 & Gene2 & Freq  & Gene1 & Gene2 & Freq  \\ \hline
    4q31.3 & 18q23 & 0.95  & 3p22.2 & 22q13.1 & 0.79  & 3q26.1 & 11p14.3 & 0.69   \\
    2p25.2 & 15q26.2 & 0.87  & 3p12.3 & 12p13.1 & 0.72  & 4q34.3 & 5q32  & 0.64   \\
    2q36.3 & 3p26.1 & 0.84  & 12q22 & 15q14 & 0.7   & 8p11.22 & 11p14.2 & 0.63   \\
    7q21.13 & 8q21.13 & 0.84  & 2p12  & Xp22.33 & 0.69  & 3q24  & 22q11.23 & 0.57   \\
    6p21.32 & 16q12.2 & 0.83  & 6p21.32 & 8p11.22 & 0.68  & 4p14  & 11p15.3 & 0.55   \\
    3p21.1 & 17p13.2 & 0.81  & 1p34.2 & 3p24.1 & 0.67  & 1q31.1 & Xq27.3 & 0.54   \\
    4q24  & 12q21.1 & 0.81  & 2p21  & Xp11.22 & 0.67  & 13q33.2 & 22q11.23 & 0.54  \\
    2q23.3 & 6p12.1 & 0.79  & 2p12  & 7p21.1 & 0.66  & 21q21.1 & 22q11.21 & 0.54   \\
    8p21.3 & 21q21.1 & 0.79  & 12q15 & 13q12.12 & 0.63  & 5q33.1 & 17q21.31 & 0.53   \\
    2q34  & 3q13.31 & 0.78  & 4q25  & 8p11.22 & 0.62  & 12q21.32 & 18q22.3 & 0.51   \\
    6p21.32 & 9q31.3 & 0.78  & 8p11.22 & Xq23  & 0.62  & 8p11.22 & 22q11.21 & 0.5    \\
    6p21.32 & 13q21.1 & 0.78  & 9p21.2 & 16q22.1 & 0.61  & 8q21.13 & Xp22.11 & 0.5    \\
    6p21.31 & 11p15.2 & 0.78  & 3p21.1 & 11q14.1 & 0.58  &       &       &       \\
    11p15.1 & 14q22.2 & 0.78  & 3p13  & 9p24.2 & 0.58  &       &       &         \\
    1p36.11 & 2p21  & 0.77  & 9q22.32 & 12p13.31 & 0.57  &       &       &       \\ \cline{7-9}
    1p31.1 & 2q32.2 & 0.76  & 7q21.3 & 22q12.3 & 0.56  & \multicolumn{3}{c}{Tumor stage}     \\ \cline{7-9}
    1q31.1 & 22q11.21 & 0.76  & 3q26.1 & 11p13 & 0.55  & Gene1 & Gene2 & Freq     \\ \cline{7-9}
    2q32.1 & 6q14.1 & 0.76  & 4q35.2 & 22q12.3 & 0.55  & 16q23.3 & 17p13.1 & 0.61      \\
    9q21.11 & 16q21 & 0.76  & 15q22.33 & 17p11.2 & 0.55  & 12p11.23 & 16q12.2 & 0.59    \\
    9q31.3 & 14q24.3 & 0.76  & 3p22.1 & 6p21.31 & 0.54  & 3q13.13 & Xq23  & 0.57   \\
    10q25.3 & 12p13.31 & 0.76  & 4q28.2 & 7q21.13 & 0.54  & 7p21.3 & 12p11.23 & 0.56      \\
    4q35.1 & 15q22.2 & 0.75  & 5q13.1 & 6q22.33 & 0.54  & 9q34.13 & 15q21.1 & 0.55      \\
    3p21.31 & 17p11.2 & 0.74  & 5q23.2 & 8p21.2 & 0.54  & 11q24.2 & 13q32.3 & 0.55     \\
    6p21.32 & 13q31.2 & 0.74  & 16q22.1 & 17q21.31 & 0.54  & 8q21.13 & 13q33.1 & 0.54          \\
    10q11.21 & 12p13.32 & 0.74  & 4q28.3 & 9p21.3 & 0.53  & 2p21  & 12p13.31 & 0.53  \\
    9q33.1 & 14q12 & 0.73  & 4q35.1 & 9p21.3 & 0.53  & 10q26.3 & 17p11.2 & 0.53   \\
    12p13.31 & 17q11.2 & 0.73  & 4q35.2 & 16q22.1 & 0.53  & 7p21.3 & 12p12.1 & 0.51     \\
    1p34.2 & 3p22.1 & 0.72  & 2q31.3 & 4q13.2 & 0.52  & 3q13.13 & 7p21.3 & 0.5     \\
    5q33.1 & 11p15.4 & 0.72  & 3p26.1 & 14q13.1 & 0.52  & 9q34.13 & 15q22.1 & 0.5      \\
    6q12  & 20p12.1 & 0.72  & 4p16.1 & 13q31.1 & 0.52  &       &       &      \\
    12p12.2 & Xp11.4 & 0.72  & 6p21.31 & 11q14.2 & 0.52  &       &       &        \\
    4q35.2 & 9p21.2 & 0.71  & 3p25.1 & 11p15.2 & 0.51  &       &       &       \\
    11p15.2 & 18q12.1 & 0.71  & 5q14.2 & Xq27.1 & 0.51  &       &       &        \\
    1p21.1 & 7q21.12 & 0.7   & 5q14.2 & Xq27.2 & 0.51  &       &       &     \\
    2p16.1 & 6p12.3 & 0.7   & 8p11.22 & 15q14 & 0.51  &       &       &        \\
    2q31.2 & 3p26.2 & 0.7   & 10q23.32 & 21q21.1 & 0.51  &       &       &     \\
    2q36.3 & 9q22.31 & 0.7   & 16q22.1 & 17p13.2 & 0.51  &       &       &        \\
    3p22.1 & 15q25.3 & 0.7   & 3p22.1 & 5q33.3 & 0.5   &       &       &      \\
    6p21.32 & Xp11.4 & 0.7   & 5q14.2 & 17q21.2 & 0.5   &       &       &      \\
    \hline
    \end{tabular}%
\end{table}%

We also searched the network for hubs (highly connected nodes), which often have important roles in genetic regulatory pathways.  Since there can be different hubs associated with different covariates, we separate them as follows.   For each node $j$, covariate $l$, and stability selection subsample $m$, let the ``covariate-specific'' degree of node $j$ be $d_{j,l}^m = \#\{k: \hat{\theta}_{jkl}\neq 0\}$.  A ranking of nodes  can then be produced for each covariate $l$ and each replication $m$, with $r_{j,l}^m$ being the corresponding rank.  Finally, we compute the median rank across all stability selection subsamples $r_{j,l} = \textrm{median}\{r_{j,l}^m, m = 1, \ldots, 100\}$, and order nodes by rank for each covariate.  The results are listed in Table \ref{tab:hubs}. Interestingly, cytoband 8p11.22 was ranked close to the top for all three covariates. The 8p11-p12 genomic region plays an important roles in breast cancer, as numerous studies have identified this region as the location of multiple oncogenes and tumor suppressor genes \citep{Yang06, Adelaide98}. High frequency of loss of heterozygosity (LOH) of this region in breast cancer has also been reported \citep{Adelaide98}. Particularly, cytoband 8p11.22 harbors the candidate tumor suppressor gene TACC1 (transforming, acidic coiled-coil containing protein 1), whose alteration is believed to disturb important regulations and participate in breast carcinogenesis \citep{Conte02}. From Table \ref{tab:results}, we can also see that the deletion of cytoband 8p11.22 region is associated with the deletion of cytoband 6p21.32 and 11p14.2 with relatively high confidence (selection frequency $>$ 0.6); and these associations change with both TP53 status and ER status. This finding is interesting because high frequency LOH at 6q and 11p in breast cancer cells are among the earliest findings that led to the discovery of recessive tumor suppressor genes of breast cancer \citep{Ali87, Devilce91, Negrini94}. Moreover, there is evidence that allele loss of c-Ha-ras locus at 11p14 correlates with paucity of oestrogen receptor protein, as well as patient survival \citep{Mackay88, Garcia89}.   These results together with the associations we detected confirm the likely cooperative roles of multiple tumor suppressor genes involved in breast cancer.

\begin{table}[htbp]\footnotesize
  \centering
  \caption{Degree-based ranking of nodes }
\label{tab:hubs}
    \begin{tabular}{ll | ll | ll | ll }
    \hline
     \multicolumn{2}{c|}{Main effect} & \multicolumn{2}{c|}{TP53 mutation status} & \multicolumn{2}{c|}{ER status} & \multicolumn{2}{c}{Tumor stage}\\     \hline
    Gene  & Median rank & Gene  & Median rank & Gene  & Median rank & Gene  & Median rank \\ \hline
    1p36.11 & 16.75 & 8p11.22 & 12.75 & 3q26.1 & 10    & 16q23.1 & 19.25 \\
    1q31.1 & 21    & 1p31.3 & 14.5  & 1q31.1 & 12    & 10q11.23 & 22.25 \\
    6p21.31 & 24.25 & 3p22.2 & 25.25 & 3p22.2 & 13    & 16q12.2 & 23.5 \\
    6p21.32 & 37    & 1q31.1 & 28.75 & 8q21.13 & 14    & 9q34.13 & 27.5 \\
    2p12  & 38.5  & 12q23.1 & 32    & 10q22.1 & 15.25 & 22q11.23 & 27.75 \\
    2q32.2 & 43    & 2p16.2 & 33.5  & 8p11.22 & 19    & 12p11.23 & 33 \\
    8q21.13 & 44.5  & 4q31.1 & 41.75 & 3p21.1 & 20.25 & 2q33.1 & 35.25 \\
    6p12.3 & 45.5  & 9p21.3 & 42    & 11q23.3 & 22    & 8p11.22 & 35.75 \\
    2q32.3 & 53.75 & 7q21.3 & 44.25 & 5q13.1 & 28    & 10q25.2 & 36 \\
    3p22.2 & 54.25 & 3q26.1 & 44.75 & 4p16.1 & 33    & 11q14.1 & 40.5 \\
    6p12.1 & 57.5  & 12q15 & 45.5  & 5q13.3 & 34    & 10p12.2 & 41.5 \\
    1p31.3 & 59.25 & 12p11.22 & 51.5  & 9p22.3 & 36.25 & 3q13.13 & 42 \\
    21q21.1 & 60    & 15q22.1 & 51.5  & 8p21.3 & 41.25 & 13q13.2 & 42.75 \\
    3q26.1 & 73.25 & 15q23 & 51.75 & 3p25.1 & 42.5  & 16q12.1 & 47 \\
    12p11.22 & 73.25 & 8q21.13 & 54    & 10q23.2 & 42.75 & 6p21.31 & 50 \\
    6q26  & 74.5  & 9p21.2 & 54.5  & 5q32  & 47    & 11q22.2 & 53 \\
    13q32.1 & 75.75 & 21q21.1 & 55.25 & 1p36.11 & 47.5  & 10q26.3 & 53.5 \\
    17p13.2 & 78    & 9q34.13 & 59    & Xp22.22 & 48.75 & 9q33.1 & 55.5 \\
    11q14.1 & 80.25 & 9p24.2 & 62    & 21q21.1 & 49    & 4q21.1 & 56 \\
    \hline
    \end{tabular}%
  \label{tab:addlabel}%
\end{table}%

\section{Summary and Discussion}
\label{sec:discuss}
We have proposed a novel Ising graphical model which allows us to incorporate extraneous factors into the graphical model in the form of covariates. Including covariates into the model allows for subject-specific graphical models, where the strength of association between nodes varies smoothly with the values of covariates.  One consequence of this is that if all covariates are continuous, there is probability 0 of the graph structure changing with covariates, and only the strength of the links is affected.   With binary covariates, which is the case in our motivating application, this situation does not arise, but in principle this could be seen as a limitation.   On the other hand, this is a necessary consequence of continuity, and small changes in the covariates resulting in large changes in the graph, as can happen with the approach of \citet{Liu10}, make the model  interpretation difficult.  Further, our approach has the additional advantage of discovering exactly which covariates affect which edges, which can be more important in terms of scientific insight.    

While here we focused on binary network data, the idea can be easily extended to categorical and Gaussian data, and to mixed graphical models involving both discrete and continuous data.  Another direction of interest is understanding conditions under which methods based on the neighborhood selection principle of running separate regressions are preferable to pseudo-likelihood type methods, and vice versa.  This comparison arises frequently in the literature, and understanding this general principle would have applications far beyond our particular method.  

\section*{Acknowledgements}
E. Levina's research is partially supported by NSF grants DMS-1106772 and DMS-1159005 and NIH grant 5-R01-AR-056646-03.  J. Zhu's research is partially supported by NSF grant DMS-0748389 and NIH grant R01GM096194.

\bibliographystyle{biometrika}
\bibliography{allref}		

\section*{Appendix: Proof of Theorem \ref{thm:main}}
\noindent For notational convenience, we omit the $j$ indexing each separate regression.   Following the literature, we prove the main theorem in two steps:   first, we prove the result holds when assumptions \textbf{A1} and \textbf{A2} hold for $\V{I}^n$ and $\V{U}^n$, the sample versions of of  $\V{I}^*$ and $\V{U}^*$ defined in \eqref{def_IU} (Proposition \ref{prop:sample_version}).     Then we show that if \textbf{A1} and \textbf{A2} hold for the population versions $\V{I}^*$ and $\V{U}^*$, they also hold for $\V{I}^n$ and $\V{U}^n$ with high probability (Proposition \ref{prop:condition_consistency}).  The sample quantities $\V{I}^n$ and $\V{U}^n$ are defined as   
\begin{eqnarray*} 
\V{I}^n &=& \nabla^2 \ell(\V{\theta}^*, \mathcal{D}_n) = \frac{1}{n} \displaystyle \sum_{i=1}^n \left(p_j^i(1 - p_j^i)(\V{x}^i \otimes \V{y}^i_{\backslash j})(\V{x}^i \otimes \V{y}^i_{\backslash j})^T\right) \ , \\
\V{U}^n &=& \frac{1}{n} \displaystyle \sum_{i=1}^n (\V{x}^i \otimes \V{y}^i_{\backslash j}) (\V{x}^i \otimes \V{y}^i_{\backslash j})^T \ . 
\end{eqnarray*}
\begin{proposition}
\label{prop:sample_version}
If $\V{A1}$ and $\V{A2}$ are satisfied by $\V{I}^n$ and $\V{U}^n$, assume moreover that 
\begin{eqnarray*}
M_n &=& \sup \lVert\V{x}\rVert_{\infty} < \infty \ \ \textrm{a.s.,}\\
\lambda_n &\geq& \frac{8M_n(2-\alpha)}{\alpha}\sqrt{\frac{\log p+\log q}{n}} \ , \\
n &>& Cd^2(\log p +\log q) \ .
\end{eqnarray*}
Then with probability at least $1 - 2 \exp\left(-C\frac{\lambda_n^2n}{M_n^2}\right)$, the result of Theorem \ref{thm:main} holds.
\end{proposition}

{\em Proof of Proposition \ref{prop:sample_version}}.  The proof requires several steps.  The uniqueness part follows directly from the following lemma:
\begin{lemma} \label{lem:uniqueness}
\emph{(Shared sparsity and uniqueness of $\hat{\V{\theta}}$, \citet{Ravikumar10})}.  
Define the sign vector $\V{t}$ for $\V{\theta}$ to satisfy the following properties,
\begin{equation*}
	 \begin{cases} \hat{t}_{k} = \mbox{sign}(\hat{\theta}_{k}), & \mbox{if } \hat{\theta}_{k} \neq 0 \ ,  \\ 
                           \rvert \hat{t}_{k} \rvert \leq 1, & \mbox{if } \hat{\theta}_{k} = 0 \ . \end{cases}
\end{equation*}
Suppose there exists an optimal solution $\hat{\V{\theta}}$ with sign $\hat{\V{t}}$ defined as above, such that, $\lVert \hat{\V{t}}_{\mathcal{S}^C}\rVert_{\infty} < 1$, then any optimal solution $\tilde{\V{\theta}}$ must have $\tilde{\V{\theta}}_{\mathcal{S}^C} = 0$. Furthermore, if the Hessian matrix $\nabla^2 \ell(\hat{\V{\theta}})_{\mathcal{SS}}$ is strictly positive definite, then $\hat{\V{\theta}}$ is the unique solution. \\
\end{lemma}

We now proceed to prove the rest of Proposition \ref{prop:sample_version}. For $\hat{\V{\theta}}$ to be a solution of \eqref{minimization}, the sub-gradient at $\hat{\V{\theta}}$ must be 0, i.e., 
 \begin{equation}\label{subgradient}
\nabla \ell(\hat{\V{\theta}}, \mathcal{D}_n) + \lambda_n \hat{\V t} = 0 \ .
\end{equation}
Then we can write   
$\nabla \ell\left(\hat{\V{\theta}}, \mathcal{D}_n\right) - \nabla \ell\left(\V{\theta}^*, \mathcal{D}_n\right) = - \lambda_n \hat{\V t} + W^n $, where 
\begin{equation*}
W^n = - \nabla \ell\left(\V{\theta}^*, \mathcal{D}_n\right) = \frac{1}{n}\sum_{i = 1}^n(\V{x}^i \otimes \V{y}_{\backslash j}^i)(y_j^i - p_j^i(\V{\theta}^*)) \ .
\end{equation*}

Let $\tilde{\V{\theta}}$ denote a point in the line segment connecting $\hat{\V{\theta}}$ and $\V{\theta}^*$.  Applying the mean value theorem gives 
\begin{equation}
\label{main_structure}
\V{I}^n\left(\hat{\V{\theta}} - \V{\theta}^*\right) = W^n - \lambda_n \hat{\V t} + R^n\ .
\end{equation}
where $R^n = \left(\nabla^2 \ell\left(\V{\theta^*}, \mathcal{D}_n\right) - \nabla^2 \ell(\tilde{\V{\theta}}, \mathcal{D}_n)\right)(\hat{\V{\theta}} - \V{\theta}^* )$.

Now define $\hat{\V{\theta}}$ as follows: let $\mathcal{S}$ be the index set of true non-zeros in $\V\theta^*$, 
let $\hat{\V{\theta}}_\mathcal{S}$ be the solution of 
\begin{equation}
\min_{(\hat{\V{\theta}}_\mathcal{S}, 0)} \ell(\hat{\V{\theta}}, \mathcal{D}_n) + \lambda_n \|\hat{\V{\theta}}_\mathcal{S}\|_1 \ ,
\end{equation}
and let $\hat{\V{\theta}}_{\mathcal{S}^C}$ = 0.    We will show that this $\hat{\V{\theta}}$ is the optimal solution and is sign consistent with high probability. 
 
We set the corresponding sign vector $\hat{\V{t}}_\mathcal{S}$ for $\hat{\V{\theta}}_\mathcal{S}$ similarly defined as in Lemma \ref{lem:uniqueness}, and $\hat{\V{t}}_{\mathcal{S}^C} = -\frac{1}{\lambda_n}\nabla_{\mathcal{S}^C} \ell(\hat{\V{\theta}}_\mathcal{S}, \mathcal{D}_n)$ as obtained in \eqref{subgradient}. Now we need to show that with high probability, 
\begin{eqnarray}
\label{tsc_toshow}
\|\hat{\V{t}_j}\|_\infty &<& 1, \ \ \ \ \ \ \  \ \ \ \ \ \textrm{for} \ \ j \in \mathcal{S}^C \\
\label{ts_toshow}
\hat{\V{t}_j} &=& sign(\V{\theta}_j^*), \ \ \textrm{for}\ \  j \in \mathcal{S} \ \textrm{and} \ \| \theta_j^* \| \geq \frac{10\lambda_n\sqrt{d}}{\Delta_{\min}}
\end{eqnarray}
The following three lemmas form the proof.  

\begin{lemma}\label{lem:control_Wn}
\emph{(Control the remainder term $W^n$)}.  
For $\alpha \in (0, 1]$, assume $\lVert\V{x}\rVert_{\infty} \leq M_n$ a.s,  then,
\begin{equation*}
P\left(\frac{2 - \alpha}{\lambda_n}\lVert W^n\rVert_{\infty} \geq \frac{\alpha}{4}\right) \leq 4 \exp \left(-\frac{\lambda_n^2n\alpha^2}{32M_n^2(2 - \alpha)^2} + \log{p} + \log{q}\right) \ .
\end{equation*}
\noindent This probability goes to 0 as long as $\lambda_n \geq 8M\frac{2 - \alpha}{\alpha} \sqrt{\frac{\log{p} + \log{q}}{n}}$.
\end{lemma}

\noindent \textit{Proof of Lemma \ref{lem:control_Wn}. }
We can write $W^n =  \frac{1}{n}\sum_{i = 1}^n(\V{x}^i \otimes \V{y}_{\backslash j}^i)(y_j^i - p_j^i(\V{\theta}^*)) = \sum_{i=1}^n Z_i$, where $Z_{ik}$ is bounded by $M_n/n$. Thus by Azuma-Hoeffding Inequality, 
\begin{eqnarray*}
P\left(\|W^n\|_\infty \geq \frac{\lambda_n \alpha}{4 (2-\alpha)}\right) &\leq& 2pqP\left(\|W^n_k\|_\infty \geq \frac{\lambda_n \alpha}{4 (2-\alpha)}\right) \notag \\
&\leq&  4 \exp \left(-\frac{\lambda_n^2n\alpha^2}{32M_n^2(2 - \alpha)^2} + \log{p} + \log{q}\right). 
\end{eqnarray*}  \hfill\(\Box\)

\begin{lemma}\label{lem:l2_consistency} \emph{($\ell_2$-consistency of the sub-vector $\hat{\V{\theta}}_{\mathcal{S}}$)}.  
If $\lambda_n d <\frac{\Delta_{\min}^2}{10\Delta_{\max}M_n}$, and, $\lVert W^n\rVert_{\infty} \leq \frac{\lambda_n}{4}$, then 
\begin{equation*}\rVert\hat{\V{\theta}}_{\mathcal{S}} - \V{\theta}_{\mathcal{S}}^* \rVert_2 \leq \frac{5\lambda_n\sqrt{d}}{\Delta_{\min}} \ .  
\end{equation*}
\end{lemma}
\noindent \textit{Proof of Lemma \ref{lem:l2_consistency}. }
Let $G(u_\mathcal{S}) = \ell(\V{\theta}_\mathcal{S}^*+u_\mathcal{S}, \mathcal{D}_n) - \ell(\V{\theta}_\mathcal{S}^*, \mathcal{D}_n) + \lambda_n(\|\V{\theta}_\mathcal{S}^*+u_\mathcal{S}\|_1 - \|\V{\theta}_\mathcal{S}^*\|_1)$ be a function $G: \mathbb{R}^d \rightarrow \mathbb{R}$. It is easy to see that $G(u_\mathcal{S})$ is convex and it achieves its minimum at $\hat{u}_\mathcal{S} = \hat{\V{\theta}}_\mathcal{S} - \V{\theta}_\mathcal{S}^*$. Moreover, $G(0) = 0$. Thus if we can show that  $G(u_\mathcal{S})$ is positive on the set $\|u_\mathcal{S}\|_2 = B$, then we will have $\hat{u}_\mathcal{S} \leq B$ due to convexity of $G(u_\mathcal{S})$. Note that 
\begin{equation*}
G(u_\mathcal{S}) = -W_\mathcal{S}^{nT} u_\mathcal{S}+u_\mathcal{S}^T\nabla^2\ell(\V{\theta}^*_\mathcal{S}+\alpha u_\mathcal{S})u_\mathcal{S} + \lambda_n(\|\V{\theta}_\mathcal{S}^*+u_\mathcal{S}\|_1 - \|\V{\theta}_\mathcal{S}^*\|_1) \ .
\end{equation*}
Further,  
\begin{eqnarray*}
|W_\mathcal{S}^{nT}u_\mathcal{S}| &\leq& \|W^n\|_\infty\|u_\mathcal{S}\|_1 \leq \frac{\lambda_n}{4}\sqrt{d}\|u_\mathcal{S}\|_2  \ , \\
\Lambda_{\min}(\nabla^2\ell(\V{\theta}^*_\mathcal{S}+\alpha u_\mathcal{S})) &\geq& \Delta_{\min} - \Delta_{\max}M_n\sqrt{d} \|u_\mathcal{S}\|_2 \ , \\
|\lambda_n(\|\V{\theta}_\mathcal{S}^*+u_\mathcal{S}\|_1 - \|\V{\theta}_\mathcal{S}^*\|_1)| &\leq& \lambda_n\sqrt{d}\|u_\mathcal{S}\|_2 \ . 
\end{eqnarray*}
Combining all of the above, we have  
\begin{equation*}
G(u_\mathcal{S}) \geq \|u_\mathcal{S}\|_2(-\Delta_{\max}M_n\sqrt{d}\|u_\mathcal{S}\|_2^2 + \Delta_{\min}\|u_\mathcal{S}\|_2 -\frac{5}{4}\lambda_n\sqrt{d}) \ . 
\end{equation*}
Easy algebra shows that if $\lambda_n d \leq \frac{\Delta_{\min}^2}{10\Delta_{\max}M_n} $  and $B = \frac{5\lambda_n\sqrt{d}}{\Delta_{\min}}$, the result follows. 
\hfill\(\Box\)

\begin{lemma} \label{lem:control_Rn}
\emph{(Control the remainder term $R^n$). }
If $\lambda_n d \leq \frac{\Delta_{\min}^2}{100M_n \Delta_{\max}}\frac{\alpha}{2 - \alpha}$,  $\lVert W^n\rVert_{\infty} \leq \frac{\lambda_n}{4}$, then 
\begin{equation*}
\frac{\lVert R^n\rVert_{\infty}}{\lambda_n} \leq \frac{25 \Delta_{\max}}{\Delta_{\min}^2}M_n\lambda_n d \leq \frac{\alpha}{4(2-\alpha)} \ . 
\end{equation*}
\end{lemma}
\vspace{0.1in}
\noindent \textit{Proof of Lemma \ref{lem:control_Rn}.}  Recall that
\begin{eqnarray*}
R^n &=& \left(\nabla^2 \ell\left(\V{\theta}^*, \mathcal{D}_n\right) - \nabla^2 \ell\left(\tilde{\V{\theta}}, \mathcal{D}_n\right)\right)\left(\hat{\V{\theta}} - \V{\theta}^*\right)\\
&=& \frac{1}{n} \displaystyle \sum_{i=1}^n \left(p_j^i(\V{\theta}^*)(1 - p_j^i(\V{\theta}^*))-p_j^i(\tilde{\V{\theta}})(1 - p_j^i(\tilde{\V{\theta}}))\right)(\V{x}^i \otimes \V{y}^i_{\backslash j})(\V{x}^i \otimes \V{y}^i_{\backslash j})^T\left(\hat{\V{\theta}} - \V{\theta}^*\right) \ .
\end{eqnarray*}
Let $\omega_j^i(\V{\theta}) = p_j^i(\V{\theta})(1 - p_j^i(\V{\theta}))$.  The $k$-th element of $R^n$ has the form
\begin{eqnarray*}
R^n_k &=& \frac{1}{n}\displaystyle \sum_{i=1}^n (\omega_j^i(\V{\theta}^*) -\omega_j^i(\tilde{\V{\theta}}))Z_k^i (\V{x}^i \otimes \V{y}^i_{\backslash j})^T\left(\hat{\V{\theta}} - \V{\theta}^*\right)\\
&=& \frac{1}{n}\displaystyle \sum_{i=1}^n\dot{\omega}_j^{i}(\bar{\V{\theta}})Z_k^i \left(\V{\theta}^*- \tilde{\V{\theta}} \right)^T (\V{x}^i \otimes \V{y}^i_{\backslash j})(\V{x}^i \otimes \V{y}^i_{\backslash j})^T\left(\hat{\V{\theta}} - \V{\theta}^*\right) \ , 
\end{eqnarray*}
where $Z_k^i = x_l^i y_m^i$, for some $(l, m)$. By \textbf{A1} and Lemma \ref{lem:l2_consistency}, we have 
\begin{equation*}
|R^n_k| \leq M_n\Delta_{\max}\|\hat{\V{\theta}} - \V{\theta}^*\|_2^2 \leq M_n\Delta_{\max}\left(\frac{5\lambda_n\sqrt{d}}{\Delta_{\min}}\right)^2 \ .  
\end{equation*}
\hfill\(\Box\)

Putting all the lemmas together, we are ready to prove Proposition \ref{prop:sample_version}. \\
\noindent \textit{Proof of Proposition \ref{prop:sample_version}. }
Set $\lambda_n = \frac{8M_n(2-\alpha)}{\alpha}\sqrt{\frac{\log{p} + \log{q}}{n}}$.  By Lemma \ref{lem:control_Wn}, we have  $\|W^n\|_\infty \leq \frac{\lambda_n \alpha}{4 (2-\alpha)} \leq \frac{\lambda_n}{4}$ with probability at least $1-4\exp(C\lambda_n^2n/M_n^2)$.
Choosing $n \geq \frac{100^2\Delta_{\max}^2(2-\alpha)^2}{\Delta_{\min}^4\alpha^2}d^2(\log p + \log q))$, we have $\lambda_n d \leq  \frac{\Delta_{\min}^2}{100M_n \Delta_{\max}}\frac{\alpha}{2 - \alpha}$, thus the conditions of  Lemmas \ref{lem:l2_consistency} and \ref{lem:control_Rn} hold.

By rewriting \eqref{main_structure} and utilizing the fact that $\hat{\V{\theta}}_{\mathcal{S}^C} = \V{\theta}_{\mathcal{S}^C}^* = 0$, we have
\begin{eqnarray}
\label{structure_I1}
\V{I}^n_{\mathcal{S}^C\mathcal{S}}(\hat{\V{\theta}}_{\mathcal{S}} - \V{\theta}_{\mathcal{S}}^*) & = & W^n_{\mathcal{S}^C} - \lambda_n\hat{\V{t}}_{\mathcal{S}^C} + R^n_{\mathcal{S}^C} \ , \\
\label{structure_I2}
\V{I}^n_{\mathcal{S}\mathcal{S}}(\hat{\V{\theta}}_{\mathcal{S}} - \V{\theta}_{\mathcal{S}}^*) & = & W^n_{\mathcal{S}} - \lambda_n\hat{\V{t}}_{\mathcal{S}} + R^n_{\mathcal{S}} \ . 
\end{eqnarray}

Since $\V{I}^n_{\mathcal{SS}}$ is invertible by assumption, combining \eqref{structure_I1} and \eqref{structure_I2} gives 
\begin{equation}
\label{t_equation}
\V{I}^n_{\mathcal{S}^C\mathcal{S}}(\V{I}^n_{\mathcal{S}\mathcal{S}})^{-1}(W^n_{\mathcal{S}} - \lambda_n\hat{\V{t}}_{\mathcal{S}} + R^n_{\mathcal{S}}) = W^n_{\mathcal{S}^C} - \lambda_n\hat{\V{t}}_{\mathcal{S}^C} + R^n_{\mathcal{S}^C} \ . 
\end{equation} 

To show \eqref{tsc_toshow}, we reorganize \eqref{t_equation} and use results from Lemmas \ref{lem:control_Wn} and \ref{lem:control_Rn}: 
\begin{eqnarray*}
\lambda_n \|\hat{\V{t}}_{\mathcal{S}^C}\|_\infty &=& \| \V{I}^n_{\mathcal{S}^C\mathcal{S}}(\V{I}^n_{\mathcal{S}\mathcal{S}})^{-1}(W^n_{\mathcal{S}} - \lambda_n\hat{\V{t}}_{\mathcal{S}} + R^n_{\mathcal{S}}) - W^n_{\mathcal{S}^C} - R^n_{\mathcal{S}^C} \|_\infty \\
&\leq& \|\V{I}^n_{\mathcal{S}^C\mathcal{S}}(\V{I}^n_{\mathcal{S}\mathcal{S}})^{-1}\|_\infty(\|W^n\|_\infty + \lambda_n + \|R^n\|_\infty) + \|W^n\|_\infty + \|R^n\|_\infty \\
&\leq& \lambda_n (1-\frac{\alpha}{2}) \ .
\end{eqnarray*}

To show \eqref{ts_toshow}, it suffices to show that $\rVert\hat{\V{\theta}}_{\mathcal{S}} - \V{\theta}_{\mathcal{S}}^* \rVert_\infty \leq \frac{\V{\theta}_{\min}^*}{2}$. By Lemma \ref{lem:l2_consistency}, 
\begin{equation*}
\rVert\hat{\V{\theta}}_{\mathcal{S}} - \V{\theta}_{\mathcal{S}}^* \rVert_\infty \leq \frac{5\lambda_n\sqrt{d}}{\Delta_{\min}} \leq \frac{\V{\theta}_{\min}^*}{2} \ .
\end{equation*}
The last inequality follows as long as $\V{\theta}_{\min}^* \geq \frac{10\lambda_n\sqrt{d}}{\Delta_{\min}}$.  This completes the proof of Proposition \ref{prop:sample_version}.
\hfill\(\Box\)

\begin{proposition} \label{prop:condition_consistency}
 If $\V{I}^*$ and $\V{U}^*$ satisfy $\V{A1}$ and $\V{A2}$, and  $M_n = \sup \lVert\V{x}\rVert_{\infty} < \infty \ \ \textrm{a.s.}$, the following hold for any $\delta > 0$. A and B are some positive constants.
\begin{eqnarray*}
P\left\{\Lambda_{\max}\left(\frac{1}{n}\sum_{i = 1}^n(\V{x}^i \otimes \V{y}_{\backslash j}^i)(\V{x}^i \otimes \V{y}_{\backslash j}^i)^T\right) \geq D_{\max} + \delta\right\} &\leq& 2 \exp\left(-A \frac{\delta^2n}{M_n^2d^2} + B (\log p + \log q)\right) \notag \\
\\
P\left(\Lambda_{\min}(\V{I}_{\mathcal{SS}}^n) \leq C_{\min} - \delta\right) &\leq& 2 \exp\left(-A \frac{\delta^2n}{M_n^2d^2} + B \log d\right) \\
P\left(\lVert\lvert\V{I}_{\mathcal{S}^c\mathcal{S}}^n \left(\V{I}_{\mathcal{SS}}^n\right)^{-1}\rvert\rVert_{\infty} \geq 1 - \frac{\alpha}{2}\right) &\leq& \exp\left(-A\frac{n}{M_n^2d^3} + B(\log{p} + \log{q}) \right)
\end{eqnarray*}
\end{proposition}
We omit the proof of Proposition \ref{prop:condition_consistency}, which is very similar to Lemmas 5 and 6 in \citet{Ravikumar10}.  \\

\noindent  {\em Proof of Theorem \ref{thm:main}.}
With Propositions \ref{prop:sample_version} and \ref{prop:condition_consistency}, the proof of Theorem \ref{thm:main} is straightforward. Given that A1 and A2 are satisfied by $\V{I}^*$ and $\V{U}^*$ and that conditions \eqref{condition_lambda} and \eqref{condition_n} hold, on the set $\mathcal{A} = \{\V{x}: M_n = \sup \|\V{x}\| < \infty \}$   the assumptions in Proposition \ref{prop:condition_consistency} are satisfied. Thus with probability at least $1 - \exp(-\frac{C\lambda_n^2n}{M_n^2})$, the conditions of Proposition \ref{prop:sample_version} hold, and therefore the results in Theorem \ref{thm:main} hold. Finally, let $\mathcal{T}$ stand for the set where the results of Theorem \ref{thm:main} hold.  Then by \eqref{condition_xbound} and \eqref{condition_M}, we have 
\begin{equation*}
P(\mathcal{T}^c) \leq  P(\mathcal{T}^c\mid\mathcal{A}) + P(\mathcal{A}^c) \leq\exp(-\frac{C\lambda_n^2n}{M_n^2}) + \exp(-M_n^\delta) \leq \exp-(C'\lambda_n^2n)^{\delta^*}, \textrm{where} \  0 < \delta^* < 1. 
\end{equation*}
\hfill\(\Box\)

\end{document}